\documentclass[conference]{IEEEtran}

\usepackage{amsmath,amsfonts}
\usepackage{graphicx}
\usepackage{subfig}
\usepackage{listings}
\usepackage{textcomp}
\usepackage{multirow}
\usepackage{array}
\usepackage{cite}
\usepackage{url}
\usepackage{balance}
\usepackage{amsmath}
\usepackage[pages=some]{background}
\usepackage[boxruled]{algorithm2e}
\def\tablename{Table}

\newcolumntype{L}[1]{>{\raggedright\let\newline\\\arraybackslash\hspace{0pt}}m{#1}}
\newcolumntype{C}[1]{>{\centering\let\newline\\\arraybackslash\hspace{0pt}}m{#1}}
\newcolumntype{R}[1]{>{\raggedleft\let\newline\\\arraybackslash\hspace{0pt}}m{#1}}

\DeclareGraphicsExtensions{.pdf,.jpeg,.png,.jpg,.eps}

\begin{document}
\title{Sparse Deep Neural Network Graph Challenge}

\author{\IEEEauthorblockN{Jeremy Kepner$^{1,2,3}$, Simon Alford$^2$, Vijay Gadepally$^{1,2}$, \\ Michael Jones$^1$, Lauren Milechin$^4$, Ryan Robinett$^3$, Sid Samsi$^1$
\\
\IEEEauthorblockA{$^1$MIT Lincoln Laboratory Supercomputing Center, $^2$MIT Computer Science \& AI Laboratory, \\\ $^3$MIT Mathematics Deparment, $^4$MIT Dept. of Earth, Atmospheric, \& Planetary Sciences
}}}

% make the title area
\maketitle

\begin{abstract}
The MIT/IEEE/Amazon GraphChallenge.org encourages community approaches to developing new solutions for analyzing graphs and sparse data.  Sparse AI analytics present unique scalability difficulties.  The proposed Sparse Deep Neural Network (DNN) Challenge draws upon prior challenges from machine learning, high performance computing, and visual analytics to create a challenge that is reflective of emerging sparse AI systems.  The Sparse DNN Challenge is based on a mathematically well-defined DNN inference  computation and can be implemented in any programming environment. Sparse DNN inference  is amenable to both  vertex-centric implementations and array-based implementations (e.g., using the GraphBLAS.org standard).  The computations are simple enough that performance predictions can be made based on simple computing hardware models.  The input data sets are derived from the MNIST handwritten letters.   The surrounding I/O and verification provide the context for each sparse DNN inference that allows rigorous definition of both the input and the output.  Furthermore, since the proposed sparse DNN challenge is scalable in both problem size and hardware, it can be used to measure and quantitatively compare a wide range of present day and future systems.  Reference  implementations have been implemented and their serial and parallel performance have been measured.  Specifications, data, and software are publicly available at GraphChallenge.org.
\end{abstract}

% no keywords

%\IEEEpeerreviewmaketitle

\section{Introduction}
\let\thefootnote\relax\footnotetext{This material is based in part upon work supported by the NSF under grants DMS-1312831 and CCF-1533644, and USD(R\&E) under contract FA8702-15-D-0001.  Any opinions, findings, and conclusions or recommendations expressed in this material are those of the authors and do not necessarily reflect the views of the NSF or USD(R\&E).}

MIT/IEEE/Amazon GraphChallenge.org encourages community approaches to developing new solutions for analyzing graphs and sparse data.  GraphChallenge.org provides a well-defined community venue for stimulating research and highlighting innovations in graph and sparse data analysis software, hardware, algorithms, and systems. The target audience for these challenges any individual or team that seeks to highlight their contributions to graph and sparse data analysis software, hardware, algorithms, and/or systems.

As research in artificial neural networks progresses, the sizes of state-of-the-art deep neural network (DNN) architectures put increasing strain on the hardware needed to implement them \cite{7298594, kepner_exact}. In the interest of reduced storage and runtime costs, much research over the past decade has focused on the sparsification of artificial neural networks \cite{lecun1990optimal,hassibi1993second,srivastava2014dropout,iandola2016squeezenet,DBLP:journals/corr/SrinivasB15,DBLP:journals/corr/HanMD15,7298681,KepnerGilbert2011,kepner2017enabling,kumar2018ibm,kepner2018mathematics}. In the listed resources alone, the methodology of sparsification includes Hessian-based pruning \cite{lecun1990optimal,hassibi1993second}, Hebbian pruning  \cite{srivastava2014dropout}, matrix decomposition \cite{7298681}, and graph techniques \cite{kumar2018ibm,KepnerGilbert2011,kepner2017enabling,kepner2018mathematics}.

Increasingly, large amounts of data are collected from social media, sensor feeds (e.g. cameras), and scientific instruments and are being analyzed with graph and sparse data analytics to reveal the complex relationships between different data feeds \cite{darpahive}. Many graph and sparse data analytics are executed in large data centers on large cached or static data sets. The processing required is a function of both the size of the  and type of data being processed. There is also an increasing need to make decisions in real-time to understand how relationships represented in graphs or sparse data evolve. Previous research on streaming  analytics has been limited by the amount of processing required. Graph and sparse analytic updates must be performed at the speed of the incoming data. Sparseness can make the application of analytics on current processors extremely inefficient. This inefficiency has either limited the size of the data that can be addressed to only what can be held in main memory or requires an extremely large cluster of computers to make up for this inefficiency. The development of a novel sparse AI analytics system has the potential to enable the discovery of relationships as they unfold in the field rather than relying on forensic analysis in data centers. Furthermore, data scientists can explore associations previously thought impractical due to the amount of processing required.

The Subgraph Isomorphism Graph Challenge \cite{samsi2017static} and the Stochastic Block Partition Challenge~\cite{ed} have enabled a new generation of graph analysis systems by highlighting the benefits of novel innovations in these systems.  Similarly, the proposed Sparse DNN Challenge seeks to highlight innovations that are applicable to emerging sparse AI and machine learning. 

Challenges such as YOHO~\cite{yoho}, MNIST~\cite{mnist}, HPC Challenge~\cite{hpcc}, ImageNet~\cite{imagenet} and VAST~\cite{vast1,vast2} have played important roles in driving progress in fields as diverse as machine learning, high performance computing and visual analytics. YOHO is the Linguistic Data Consortium database for voice verification systems and has been a critical enabler of speech research. The MNIST database of handwritten letters has been a bedrock of the computer vision research community for two decades. HPC Challenge has been used by the supercomputing community to benchmark and acceptance test the largest systems in the world as well as stimulate research on the new parallel programing environments. ImageNet populated an image dataset according to the WordNet hierarchy consisting of over 100,000 meaningful concepts (called synonym sets or synsets)~\cite{imagenet} with an average of 1000 images per synset and has become a critical enabler of vision research. The VAST Challenge is an annual visual analytics challenge that has been held every year since 2006; each year, VAST offers a new topic and submissions are processed like conference papers.  The  Sparse DNN   Graph Challenge seeks to draw on the best of these challenges, but particularly the VAST Challenge in order to highlight innovations across the algorithms, software, hardware, and systems spectrum.

The focus on graph analytics allows the Sparse DNN  Graph Challenge to also
draw upon significant work from the graph benchmarking community.  The Graph500
(Graph500.org) benchmark (based on \cite{bader2006designing}) provides a scalable
power-law graph generator \cite{leskovec2005realistic} (used to build the world's largest synthetic graphs) with the goal of optimizing the rate of building a tree of the graph.  The
Firehose benchmark (see http://firehose.sandia.gov) simulates computer network traffic for
performing real-time analytics on network traffic.  The PageRank Pipeline benchmark
\cite{dreher2016pagerank,bisson2016cuda} uses the Graph500 generator (or any other
graph) and provides reference implementations in multiple programming languages to allow users to
optimize the rate of computing PageRank (1st eigenvector) on a graph. Finally,
miniTri (see mantevo.org) \cite{wolf2015,wolf2016} takes an arbitrary graph as input
and optimizes the time to count triangles.

The organization of the rest of this paper is as follows.  Section II provides information on the relevant DNN mathematics and computations.  Section III describes the synthetic DNNs used in the Challenge. Section IV provides the specifics of the input feature dataset based on  MNIST images.  Section V lays out the Sparse DNN Challenge steps and example code.  Section VI discusses relevant metrics for the Challenge.  Section VII summarizes the work and describes future directions.

\section{Deep Neural Networks}

Machine learning has been the foundation of artificial intelligence since its inception
\cite{ware1955introduction,clark1955generalization,selfridge1955pattern,dinneen1955programming,newell1955chess,mccarthy2006proposal,minsky1960learning,minsky1961steps}. Standard machine learning applications include speech recognition \cite{selfridge1955pattern}, computer vision \cite{dinneen1955programming}, and even board games \cite{newell1955chess,samuel1959some}.

\begin{figure}[htb]
  	\centering
    	\includegraphics[width=\columnwidth]{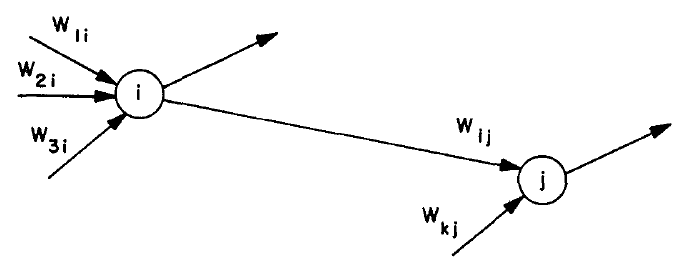}
      	\caption{Typical network elements $i$ and $j$ showing connection weights $w$ (reproduced from  \cite{clark1955generalization})}
      	\label{fig:clark1955fig1}
\end{figure}

\begin{figure}[htb]
  	\centering
    	\includegraphics[width=\columnwidth]{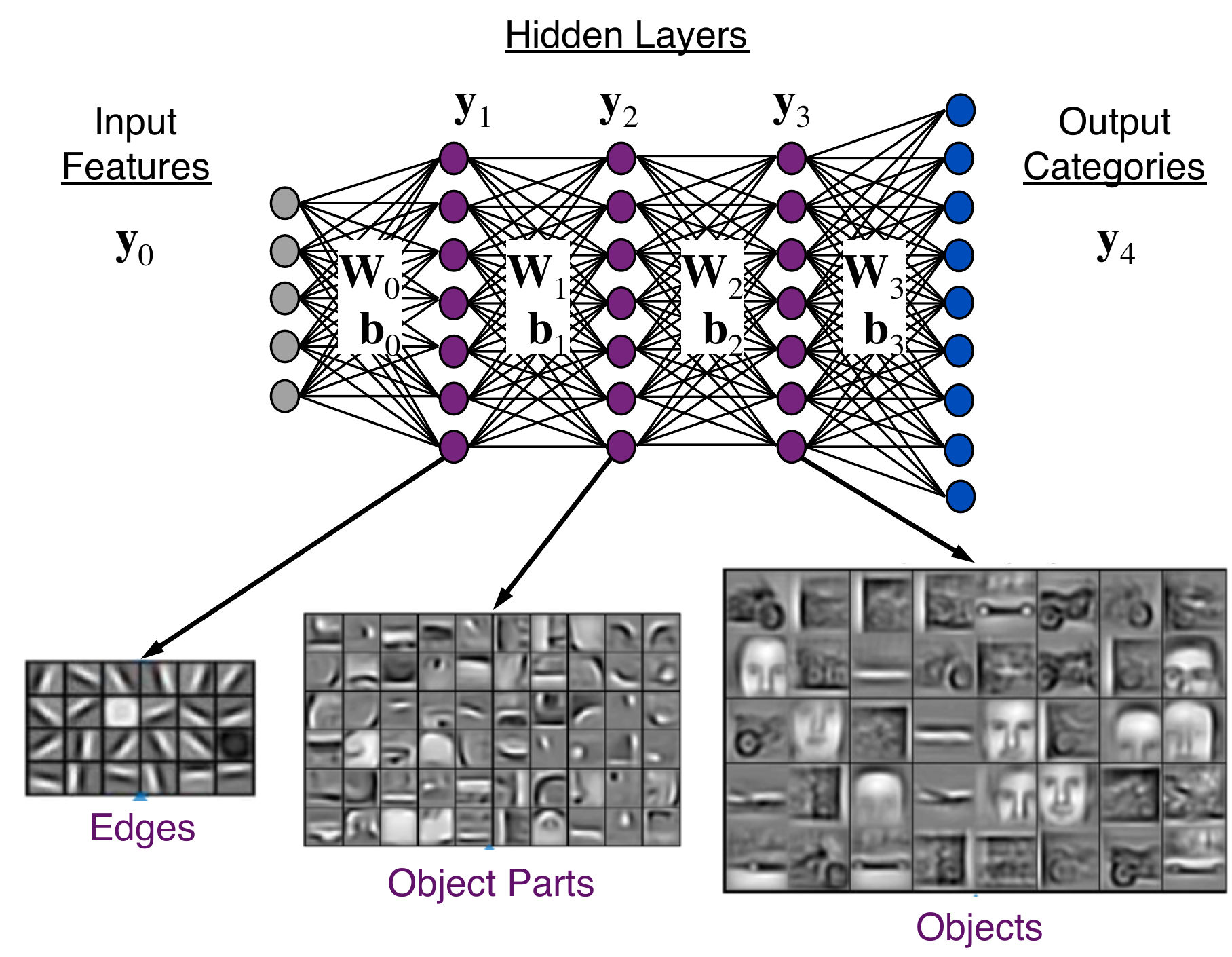}
	\caption{Four layer ($L=4$) deep neural network architecture
	for categorizing images.  The input
	features  ${\bf y}_0$ of an image are passed through a series
	of network layers ${\bf W}_{\ell=0,1,2,3}$, with bias terms
	${\bf b}_{\ell=0,1,2,3}$, that produce scores for categories
	${\bf y}_{L=4}$.  (Figure adapted from \cite{lee2009convolutional})}
      	\label{fig:DNNarchitecture}
\end{figure}

Drawing inspiration from biological neurons to implement machine learning was the topic of the first paper presented at the first machine learning conference in 1955 \cite{ware1955introduction,clark1955generalization} (see Figure~\ref{fig:clark1955fig1}). It was recognized very early on in the field that direct computational training of neural networks was computationally unfeasible with the computers that were available at that time \cite{minsky1960learning}.  The many-fold improvement in neural network computation and theory has made it possible to create neural networks capable of better-than-human performance in a variety of domains \cite{lippmann1987introduction,reynolds2000speaker,krizhevsky2012imagenet,lecun2015deep}. The production of validated data sets \cite{campbell1995testing,lecun1998mnist,deng2009imagenet} and the power of graphic processing units (GPUs) \cite{campbell2002deep,mcgraw2007benchmarking,kerr2008gpu,epstein2012making}
have allowed the effective training of deep neural networks (DNNs) with 100,000s of input features, $N$, and 100s of layers, $L$, that are capable of choosing from among 100,000s categories, $M$ (see Figure~\ref{fig:DNNarchitecture}).

The impressive performance of large DNNs provides motivation to explore even larger networks.  However, increasing $N$, $L$, and $M$ each by a factor 10 results in a 1000-fold increase in the memory required for a DNN.  Because of these memory constraints, trade-offs are currently being made in terms of precision and accuracy to save storage and computation \cite{liu2015sparse,lavin2016fast,jouppi2017datacenter,kepner2017enabling}. Thus, there is significant interest in exploring the effectiveness of sparse DNN representations where many of the weight values are zero.  As a comparison, the human brain has approximately 86 billion neurons and 150 trillion synapses~\cite{CNE:CNE21974}.  Its graph representation would have approximately 2,000 edges per node, or a density of $2 \times 10^3 / 86 \times 10^9 = 0.000002\%$.

If a large fraction of the DNN weights can be set to zero, storage and computation costs can be reduced proportionately \cite{iandola2016squeezenet,shi2017speeding}.  The interest in sparse DNNs is not limited to their computational advantages. There has also been extensive theoretical work exploring the potential neuromorphic and algorithmic benefits of sparsity \cite{lee2008sparse,boureau2008sparse,glorot2011deep,DBLP:journals/corr/HanMD15,yu2012exploiting}.

  The primary mathematical operation performed by a DNN network is the inference, or forward propagation, step.  Inference is executed repeatedly during training to determine both the weight matrix ${\bf W}_\ell$ and the bias vectors ${\bf b}_\ell$ of the DNN.  The inference computation shown in Figure~\ref{fig:DNNarchitecture} is given by
$$
  {\bf y}_{\ell + 1} = h({\bf y}_\ell {\bf W}_\ell  + {\bf b}_\ell)
$$
where $h()$ is a nonlinear function applied to each element of the vector.  The Sparse DNN Challenge uses the standard graph community convention whereby ${\bf W}(i,j) \neq 0$  implies a connection between neuron $i$ and neuron $j$.  In this convention ${\bf y}_\ell$ are row vectors and left matrix multiply  is used to progress through the network.  Standard AI definitions can be used  by transposing all matrices and multiplying on the  right.
A commonly used function is the rectified linear unit (ReLU) given by
$$
   h({\bf y}) = \max({\bf y},0)
$$
which sets values less that 0 to 0 and leaves other values unchanged.  For the Sparse DNN challenge, $h()$ also has an upper limit set to 32. When training a DNN, or performing inference on many different inputs, it is usually necessary to compute multiple ${\bf y}_\ell$ vectors at once in a batch that can be denoted as the matrix ${\bf Y}_\ell$.  In matrix form, the inference step becomes
$$
  {\bf Y}_{\ell + 1} = h({\bf Y}_\ell {\bf W}_\ell  + {\bf B}_\ell)
$$
where ${\bf B}_\ell$ is a replication of ${\bf b}_\ell$ along columns given by
$$
  \mathbf{B}_\ell = \mathbf{b}_\ell |\mathbf{Y}_\ell \mathbf{1}|_0
$$
and $\mathbf{1}$ is a column array of 1's, and $| ~ |_0$ is the zero norm.

\section{Neural Network Data}

Scale is an important driver of the Graph Challenge and graphs with billions to trillions of edges are of keen interest. Real sparse neural networks of this size are difficult to obtain from real data.  Until such data is available, a reasonable first step is to simulate data with the desired network properties with an emphasis on the difficult part of the problem, in this case: large sparse DNNs.

\begin{figure}[htb]
  	\centering
    	\includegraphics[width=\columnwidth]{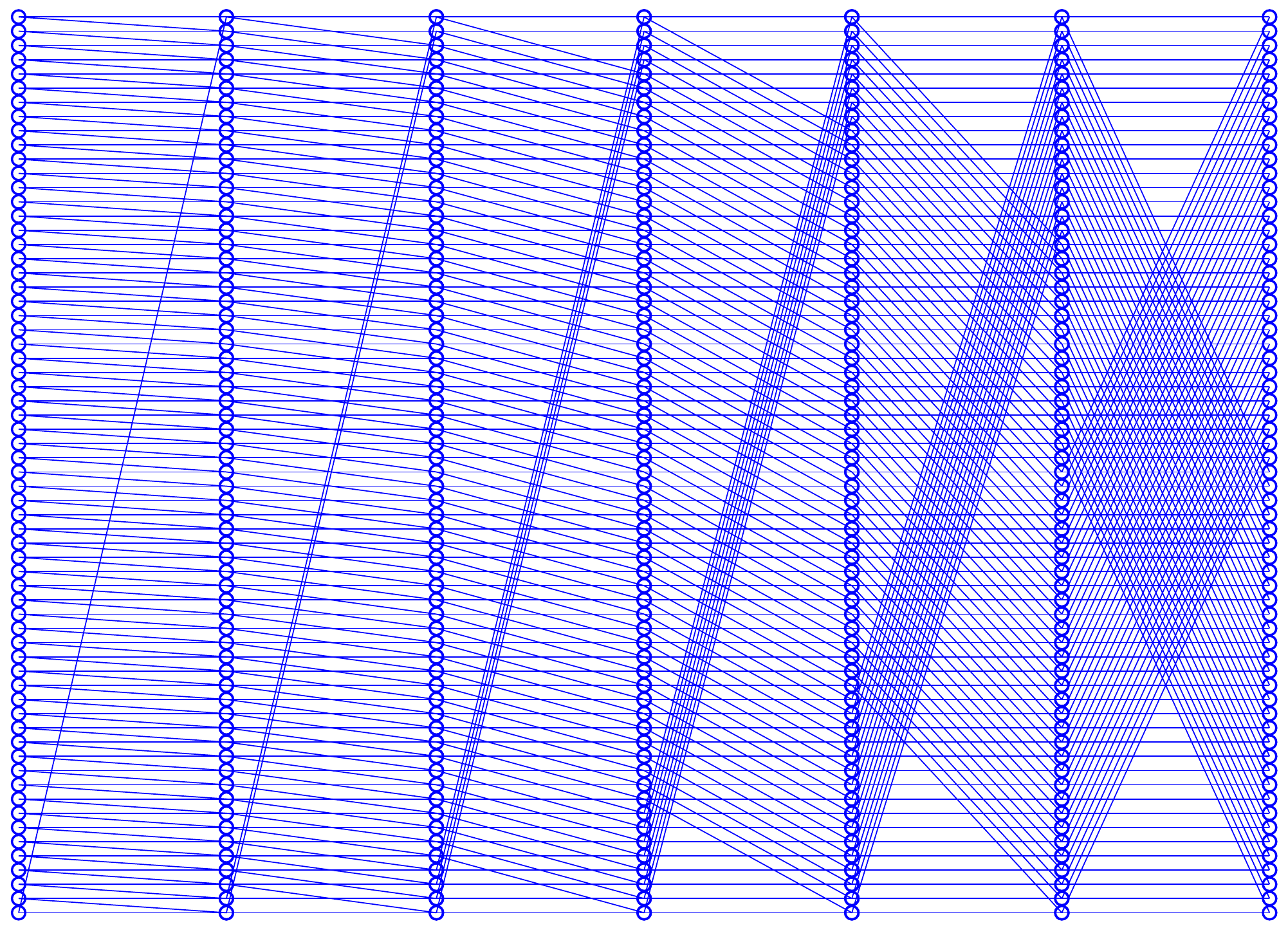}
	\caption{6 layer, 64 neurons per layer, 2 connections per neuron RadiX-Net DNN produced from Radix set [[2,2,2,2,2,2]].  The Kronecker product of this DNN with [16,16,16,16,16,16,16] produces a 6 layer, 1024 neurons per layer DNN with 32 connections per neuron.}
      	\label{fig:RadiX-Net-DNN}
\end{figure}

The RadiX-Net synthetic sparse DNN generator is used \cite{robinett2019radix-net} to efficiently generate a wide range of pre-determined DNNs.  RadiX-Net produces DNNs with a number of desirable properties, such as equal number of paths between all inputs, outputs, and intermediate layers.  The RadiX-Net DNN generation algorithm uses mixed radices to generate DNNs of specified connectedness (see Figure~\ref{fig:RadiX-Net-DNN}) which are then expanded via Kronecker products into larger DNNs.  For the Sparse DNN Challenge different DNNs were created with different numbers of neurons per layer.  The RadiX-Net parameters used to create the base DNNs are given in Table~\ref{tab:RadiX-Net-parameters}.  The base DNNs are then grown to create much deeper DNNs by repeatedly randomly permuting and appending the base DNNs.  The permutation process preserves the base DNN properties.  The scale of the resulting large sparse DNNs are shown Table~\ref{tab:RadiX-Net-DNNs}.

\begin{table*}
\centering
\begin{tabular}{cccccc}
\hline
\textbf{Radix Set} & \textbf{Kronecker Set} & \textbf{Layers} & \textbf{Neurons/Layer} & \textbf{Density} & \textbf{Bias}\\
\hline
[[2,2,2,2,2,2]]             & [16,16,16,16,16,16,16]                   & 6  & 1024  & 0.03   & -0.30 \\
\hline
[[2,2,2,2,2,2,2,2]]         & [16,16,16,16,16,16,16,16,16]             & 8  & 4096  & 0.008  & -0.35 \\
\hline
[[2,2,2,2,2,2,2,2,2,2]]     & [16,16,16,16,16,16,16,16,16,16,16]       & 10 & 16384 & 0.002  & -0.40 \\
\hline
[[2,2,2,2,2,2,2,2,2,2,2,2]] & [16,16,16,16,16,16,16,16,16,16,16,16,16] & 12 & 65536 & 0.0005 & -0.45 \\
\hline
\end{tabular}
\caption{RadiX-Net radix and Kronecker parameters and resulting base DNNs layers, neurons per layer, fraction of non-zeros in weight matrices (density), and bias value.  All DNNs have 32 connections per neurons.}
\label{tab:RadiX-Net-parameters}
\end{table*}

\begin{table}
\centering
\begin{tabular}{ccccc}
\hline
                & \textbf{Neurons} & \textbf{Neurons} & \textbf{Neurons} & \textbf{Neurons}\\
\textbf{Layers} & 1024             & 4096             & 16384            & 65536 \\
\hline
120             & 3,932,160        & 15,728,640       & 62,914,560       & 251,658,240 \\
\hline
480             & 15,728,640       & 62,914,560       & 251,658,240      & 1,006,632,960 \\
\hline
1920            & 62,914,560       & 251,658,240      & 1,006,632,960    & 4,026,531,840 \\
\hline
\end{tabular}
\caption{Total number of connections = 32x(Layers)x(Neurons) for different large sparse DNNs used in the Sparse DNN Challenge.}
\label{tab:RadiX-Net-DNNs}
\end{table}

\section{Input Data Set}

Executing the Sparse DNN Challenge requires input data or feature vectors $\mathbf{Y}_0$.  MNIST (Modified National Institute of Standards and Technology) is a large database of handwritten digits that is widely used for training and testing DNN image processing systems \cite{mnist}.  MNIST consists of 60,000 28{$\times$}28 pixel images.   The Sparse DNN Graph Challenge uses interpolated sparse versions of this entire corpus as input (Figure~\ref{fig:MNIST}).  Each 28{$\times$}28 pixel image is resized to 32{$\times$}32 (1024 neurons), 64{$\times$}64 (4096 neurons), 128{$\times$}128 (16384 neurons), and 256{$\times$}256 (65536 neurons).  The resized images are thresholded so that all values are either 0 or 1.  The images are flattened into a single row to form a feature vector.  The non-zero values are written as triples to a .tsv file where each row corresponds to a different image, each column is the non-zero pixel location and the value is 1.

\begin{figure}[htb]
  	\centering
    	\includegraphics[width=\columnwidth]{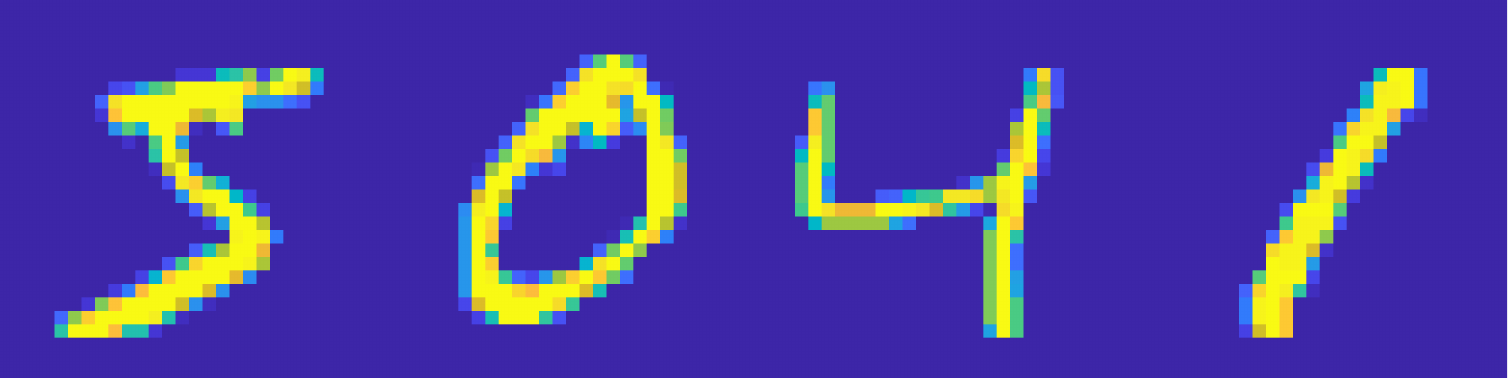}
	   	\includegraphics[width=\columnwidth]{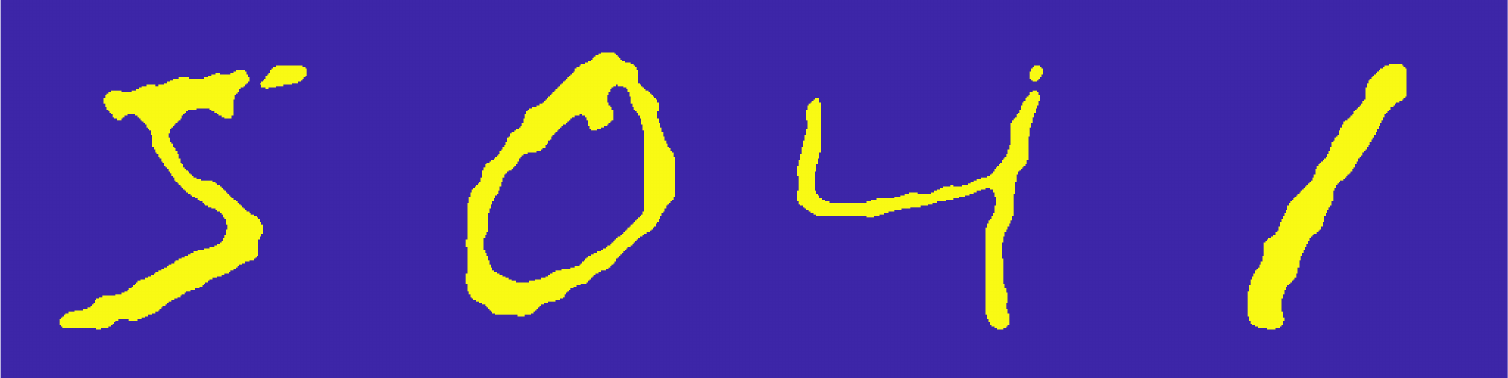}
	\caption{MNIST data set consists of 60,000 handwritten digits \cite{mnist}.  (top) Original 28{$\times$}28 pixel images of four MNIST images. (bottom) 256{$\times$}256 resampled thresholded versions of the same images.}
      	\label{fig:MNIST}
\end{figure}

\section{Sparse DNN Challenge}

The core the Sparse DNN Challenge is timing DNN inference using the provided DNNs on the provide MNIST input data and verifying the output with provided truth categories.  The complete process for performing the challenge consists of the following steps

\begin{itemize}
\item Download from GraphChallenge.org: DNN weight matrices $\mathbf{W}_\ell$, sparse MNIST input data $\mathbf{Y}_0$, and truth categories
\item Load a DNN and its corresponding input
\item Create and set the appropriate sized bias vectors $\mathbf{b}_\ell$ from the table
\item \underline{Timed}: Evaluate the DNN equation for all layers
$$
  {\bf Y}_{\ell + 1} = h({\bf Y}_\ell {\bf W}_\ell  + {\bf B}_\ell)
$$
\item \underline{Timed}: Identify the categories (rows) in final matrix with entries $> 0$
\item Compare computed categories with truth categories to check correctness
\item Compute rate for the DNN: (\# inputs) $\times$ (\# connections)  / time 
\item Report time and rate for each DNN measured
\end{itemize}
Reference serial implementations in various programming languages are available at GraphChallenge.org. The Matlab serial reference of the inference calculation is a follows

\noindent \rule{\columnwidth}{0.5pt}

{\tt
\noindent function Y = inferenceReLUvec(W,bias,Y0);

  YMAX = 32;
  
  Y = Y0;
  
  for i=1:length(W)
  
     ~ Z = Y*W\{i\};
     
     ~ b = bias\{i\};
     
     ~ Y = Z + (double(logical(Z)) .* b);
     
     ~ Y(Y < 0) = 0;
     
     ~ Y(Y > YMAX) = YMAX;
     
  end

\noindent end 
}

\noindent \rule{\columnwidth}{0.5pt}

For a given implementation of the Sparse DNN Challenge an implementor should keep the following guidance in mind.
\centerline{\underline{Do}}
\begin{itemize}
\item Use an implementation that could work on real-world data
\item Create compressed binary versions of inputs to accelerate reading the data
\item Split inputs and run in data parallel mode to achieve higher performance (this requires replicating weight matrices on every processor and can require a lot of memory)
\item Split up layers and run in a pipeline parallel mode to achieve higher performance (this saves memory, but requires communicating results after each group of layers)
\item Use other reasonable optimizations that would work on real-world data
\end{itemize}
\centerline{\underline{Avoid}}
\begin{itemize}
\item Exploiting the repetitive structure of weight matrices, weight values, and bias values
\item Exploiting layer independence of results
\item Using optimizations that would not work on real-world data
\end{itemize}

\section{Computational Metrics}
\label{sec:metrics}
Submissions to the Sparse DNN Challenge will be evaluated on the overall innovations highlighted by the implementation and two metrics: correctness and performance.

\subsection{Correctness}
\label{sec:correctness}
Correctness is evaluated by comparing the reported categories with the ground truth categories provided.

\subsection{Performance}
\label{sec:perf}
The performance of the algorithm implementation should be reported in terms of the following metrics:
\begin{itemize}
\item Total number of non-zero connections in the given DNN: This measures the amount of data processed
\item Execution time: Total time required to perform DNN inference.
\item Rate: Measures the throughput of the implementation as the ratio of the number of inputs (e.g., number of MNIST images) times the number of connections in the DNN divided by the execution time.
\item Processor: Number and type of processors used in the computation.
\end{itemize}

\subsection{Timing Measurements}
  Serial timing measurements of the Matlab code are shown in Table~\ref{tab:Timing} and provide one example for reporting results.  Parallel implementations of the Sparse DNN benchmark were developed and tested on the MIT SuperCloud TX-Green  supercomputer using pMatlab \cite{Kepner2009}.

\begin{table}
\centering
\begin{tabular}{ccccc}
\hline
\textbf{Neurons} & \textbf{Layers} & \textbf{Connections} & \textbf{Time} & \textbf{Rate}\\
\textbf{per Layer} &              & \textbf{(edges)}             & \textbf{(seconds)}            & \textbf{(inputs$\times$edges/sec)} \\
\hline
1024 & 120 & 3,932,160 & 626 & 376$\times10^6$ \\
1024 & 480 & 15,728,640 & 2440 & 386$\times10^6$ \\
1024 & 1920 & 62,914,560 & 9760 & 386$\times10^6$ \\
4096 & 120 & 15,728,640 & 2446 & 385$\times10^6$ \\
4096 & 480 & 62,914,560 & 10229 & 369$\times10^6$ \\
4096 & 1920 & 251,658,240 & 40245 & 375$\times10^6$ \\
16384 & 120 & 62,914,560 & 10956 & 344$\times10^6$ \\\
16384 & 480 & 251,658,240 & 45268 & 333$\times10^6$ \\
16384 & 1920 & 1,006,632,960 & 179401 & 336$\times10^6$ \\
65536 & 120 & 251,658,240 & 45813 & 329$\times10^6$ \\
65536 & 480 & 1,006,632,960 & 202393 & 299$\times10^6$ \\
65536 & 1920 & 4,026,531,840 & & \\
\hline
\end{tabular}
\caption{Serial timing measurements of inference rate on different sparse DNNs.}
\label{tab:Timing}
\end{table}

\begin{figure}[htb]
  	\centering
    	\includegraphics[width=\columnwidth]{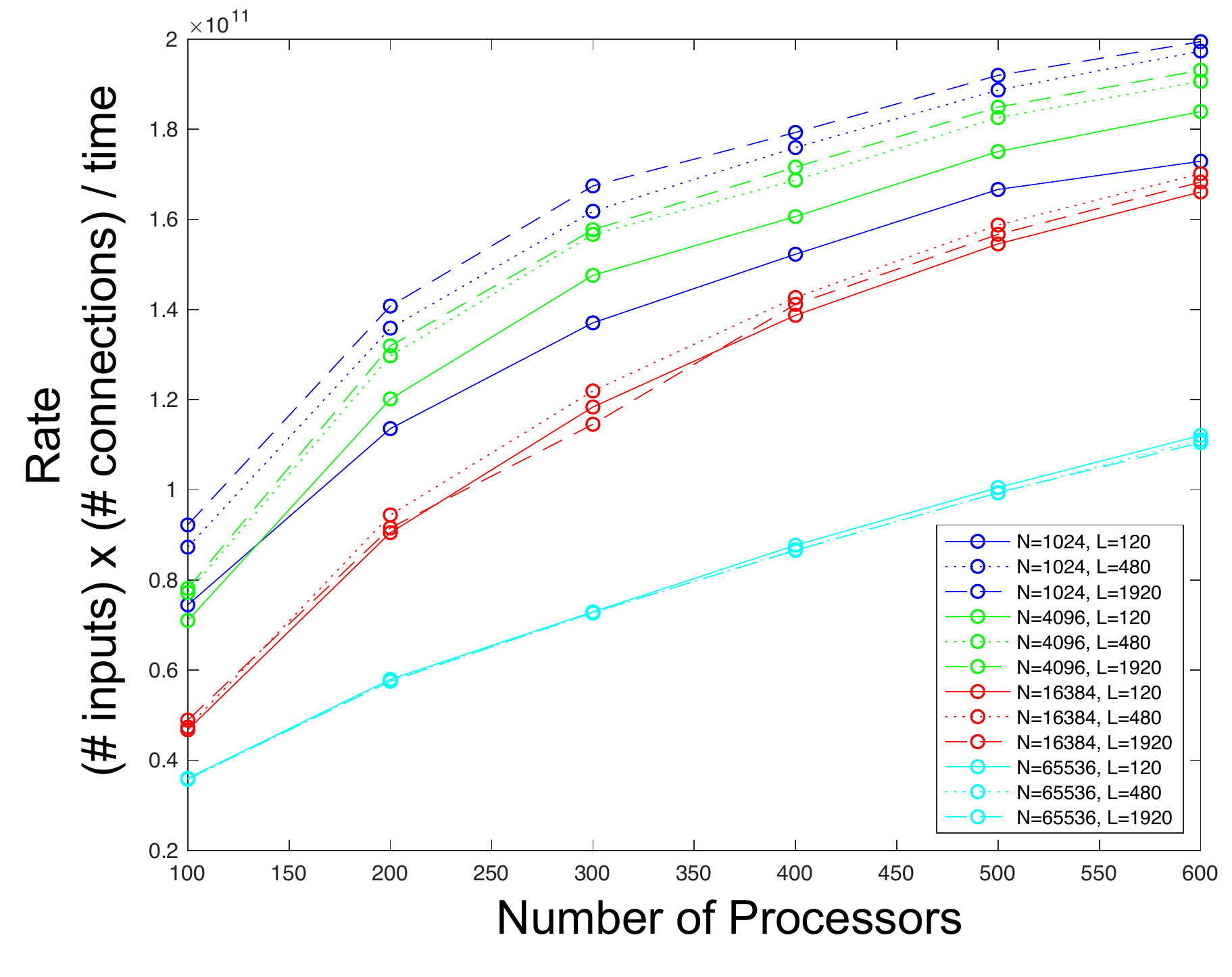}
	\caption{Inference rate versus number of processors for various DNN sizes.}
      	\label{fig:MNIST}
\end{figure}

\section{Summary}
The MIT/IEEE/Amazon GraphChallenge.org encourages community approaches to developing new solutions for analyzing graphs and sparse data.  Sparse AI analytics presents unique scalability difficulties.  The machine learning, high performance computing, and visual analytics communities have wrestled with these difficulties for decades and developed methodologies for creating challenges to move these communities forward.  The proposed Sparse Deep Neural Network (DNN) Challenge draws upon prior challenges from machine learning, high performance computing, and visual analytics to create a challenge that is reflective of emerging sparse AI systems.  The Sparse DNN Challenge is a based on a mathematically well-defined DNN inference  kernel and can be implemented in any programming environment. Sparse DNN inference  is amenable to both  vertex-centric implementations and array-based implementations (e.g., using the GraphBLAS.org standard).  The computations are simple enough that performance predictions can be made based on simple computing hardware models.  The input data sets are derived from the MNIST handwritten letters.   The surrounding I/O and verification provide the context for each sparse DNN inference that allows rigorous definition of both the input and the output.  Furthermore, since the proposed sparse DNN challenge is scalable in both problem size and hardware, it can be used to measure and quantitatively compare a wide range of present day and future systems.  Reference  implementations been implemented and their serial and parallel performance have been measured.  Specifications, data, and software are publicly available at GraphChallenge.org.

\section*{Acknowledgments}
The authors wish to acknowledge the following individuals for their contributions and support: William Arcand, David Bestor, William Bergeron, Bob Bond, Chansup Byun, Alan Edelman,  Matthew Hubbell,  Anne Klein, Charles Leiserson, Dave Martinez, Mimi McClure, Julie Mullen, Andrew Prout, Albert Reuther, Antonio Rosa, Victor Roytburd, Siddharth Samsi, Charles Yee and the entire GraphBLAS.org community for their support and helpful suggestions.

\balance

\bibliographystyle{unsrt}%{IEEEtran}
\bibliography{IEEEabrv,references.bib}

% that's all folks
\end{document}